\documentclass[sigconf]{acmart}
\AtBeginDocument{%
  }

\setcopyright{acmlicensed}
\copyrightyear{2018}
\acmYear{2018}
\acmDOI{XXXXXXX.XXXXXXX}
\acmConference[Conference acronym 'XX]{Make sure to enter the correct
  conference title from your rights confirmation email}{June 03--05,
  2018}{Woodstock, NY}
\acmISBN{978-1-4503-XXXX-X/2018/06}

\usepackage{enumitem}
\usepackage{booktabs, multirow}
\usepackage{tcolorbox}
\usepackage{graphicx}
\usepackage{subcaption}
\usepackage[table]{xcolor}
\usepackage[dvipsnames]{xcolor}
\definecolor{myG}{HTML}{D5E8D4}
\definecolor{myB}{HTML}{DAE8FC}
\definecolor{myP}{HTML}{E1D5E7}
\definecolor{myR}{HTML}{FF9999}
\begin{document}

\title[Towards Generalist Models for Molecular Property Prediction]{
Towards Generalist Models for Molecular Property Prediction: Distilling Verbalized Knowledge from Decision Trees
}


\author{Khiem Le}
\affiliation{%
\institution{Computer Science and Engineering}
\institution{University of Notre Dame, IN, USA}
\city{}\country{}}
\email{kle3@nd.edu}

\author{Marcos Martínez Galindo}
\affiliation{%
\institution{IBM Research Lab}
\institution{IBM Ireland}
\city{}\country{}}
\email{Marcos.Martinez.Galindo@ibm.com}

\author{Vanessa Lopez}
\affiliation{%
\institution{IBM Research Lab}
\institution{IBM Ireland}
\city{}\country{}}
\email{vanlopez@ie.ibm.com}

\author{Sreejata Dey}
\affiliation{%
\institution{Applied \& Computational Math. Stat.}
\institution{University of Notre Dame, IN, USA}
\city{}\country{}}
\email{sdey@nd.edu}

\author{Ting Hua}
\affiliation{%
\institution{Computer Science and Engineering}
\institution{University of Notre Dame, IN, USA}
\city{}\country{}}
\email{thua@nd.edu}

\author{Nitesh V. Chawla}
\affiliation{%
\institution{Computer Science and Engineering}
\institution{University of Notre Dame, IN, USA}
\city{}\country{}}
\email{nchawla@nd.edu}

\author{Hoang Thanh Lam}
\affiliation{%
\institution{IBM Research Lab}
\institution{IBM Ireland}
\city{}\country{}}
\email{t.l.hoang@ie.ibm.com}

\renewcommand{\shortauthors}{Khiem et al.}

\begin{abstract}
Molecular Property Prediction (MPP) is a fundamental problem in drug discovery that has recently attracted growing attention. Large Language Models (LLMs), known for their impressive proficiency across domains, show promise as generalist models for MPP. However, their current performance remains below the threshold needed for practical adoption. To bridge this gap, we propose TreeKD for distilling the knowledge of tree-based specialist models into LLMs to complement the internal knowledge of LLMs and improve their predictive accuracy. For each property, we train a specialist decision tree using features derived from 40K functional groups in the input molecules. Then, the predictive rule learned by the decision tree, which encodes its knowledge, is verbalized and incorporated into the prompts for training LLMs. In addition, by replacing a single decision tree with a Random Forest, we introduce a test-time scaling technique called rule-consistency, which aggregates predictions generated from different prompts constructed with different rules. An extensive evaluation with two LLMs, Gemma-2-2B and Granite-3.3-2B, on the TDC benchmark with 22 prediction tasks shows that our method substantially enhances the performance of LLMs, advancing the development of generalist models for MPP. 
\end{abstract}




\maketitle

\section{Introduction}

Molecular Property Prediction (MPP) plays a crucial role in the drug discovery pipeline \cite{wu2018, liyaqat2025advancements}, aiming to predict quantitative characteristics of drug candidates such as ADMET properties (Absorption, Distribution, Metabolism, Excretion, and Toxicity). Accurate prediction enables reliable screening of potential candidates at early stages, lowering experimental costs and optimizing efficiency. While traditional specialist models achieve strong performance on individual properties, they lack unified and interactive capabilities. The recent success of Large Language Models (LLMs) \cite{achiam2023gpt, comanici2025gemini} across diverse domains \cite{hendrycks2021measuring, suzgun-etal-2023-challenging} has motivated increasing efforts to develop generalist models for MPP \cite{zhao2023gimlet, zhang2025scientific}. Unlike specialist models, generalist models can handle multiple properties within a single interface, enabling streamlined and autonomous workflows. Moreover, they allow users to interact with them via natural language, obtain explanations for predictions, and engage in scientific conversations \cite{zeng2024chatmol, de2025multimodal}. Such capabilities may also lower barriers for non-expert users \cite{2025-interact}. However, their current performance remains insufficient for practical adoption \cite{guo2023can, zhong2024benchmarking}. 

\begin{figure}
\centering
\includegraphics[width=1\linewidth]{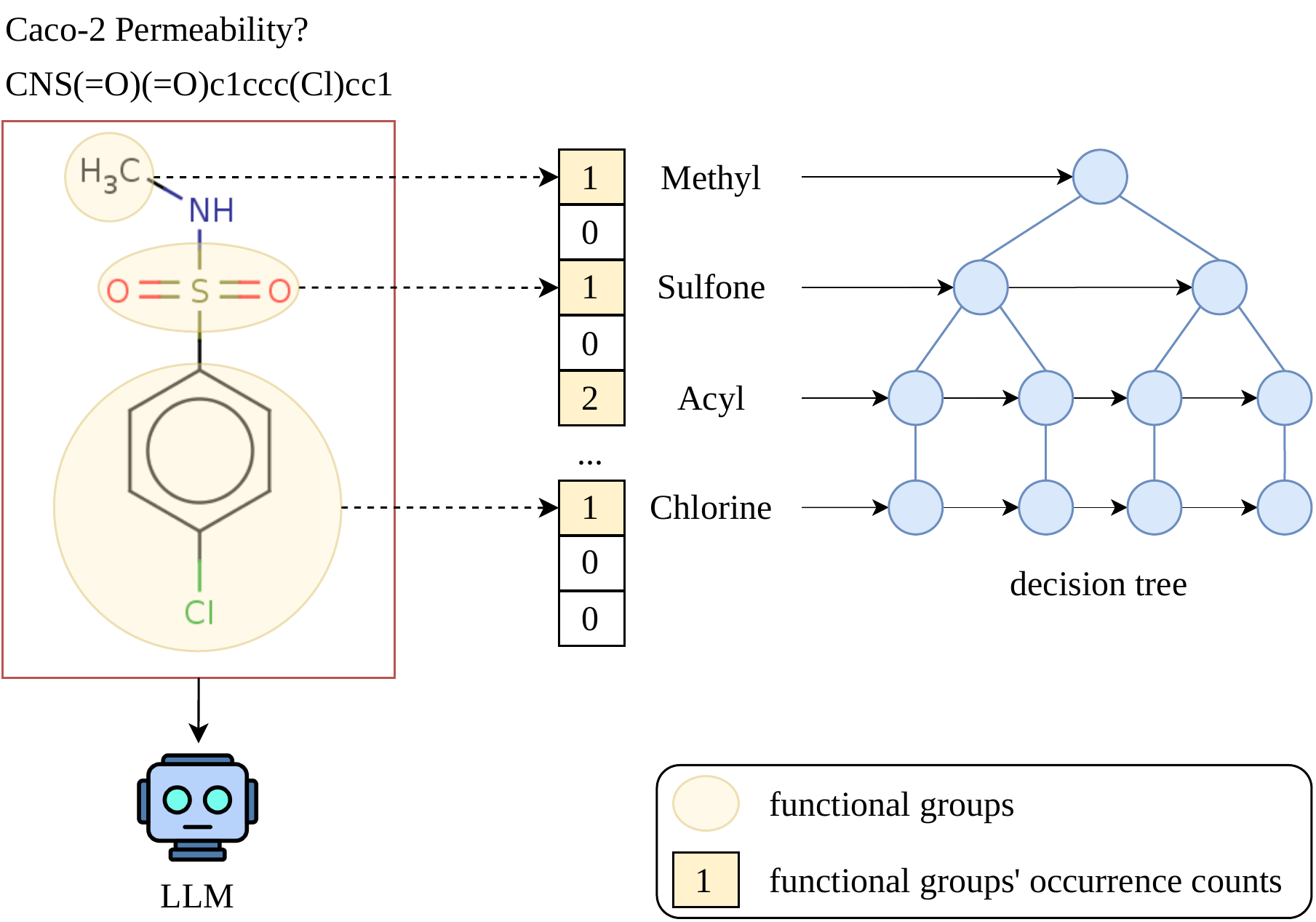}
\caption{
An example illustrates that the LLM sees a molecule as a whole, while a decision tree operates on each dimension of functional groups. 
}
\label{fig-1}
\end{figure}

A key reason behind this performance gap is that LLMs tend to attend to molecules as a whole given their SMILES string (globally), while struggling to recognize and utilize their local substructures \cite{zhao2023scientific, xia2023understanding, runcie2025assessing} (illustrated in Figure \ref{fig-1}). Yet, these substructures are the most informative for molecules and largely determine their properties \cite{rogers2010extended, mauri2017molecular}. In contrast, tree-based specialist models are well-suited for molecules because they learn to make predictions based on individual dimensions of local substructure features separately. In addition, due to this characteristic, they are more sensitive to the concept of activity cliffs, i.e., situations where small structural changes can lead to large shifts in properties \cite{maggiora2006outliers, stumpfe2012exploring}. Motivated by these insights, we ask: \textit{Can we harness the knowledge of tree-based specialist models on local substructures to assist LLMs in MPP?}

In this work, we propose TreeKD, a novel knowledge distillation method that incorporates the verbalized knowledge of tree-based specialist models into LLMs to complement their internal knowledge and improve their predictive accuracy. First, we extract all functional groups (FGs) present in the input molecules, where FGs are predefined local substructures that can be described using standardized names (e.g., Methyl, Sulfone, Acyl, Chlorine, …) \cite{gutermuth2025smartchemist}. We explicitly provide LLMs with identified FGs via standardized names, rather than relying on implicit recognition. This reduces the semantic gap between the SMILES string and natural language \cite{wellawatte2025human}, the modality LLMs are pre-trained on. Beyond that, we use the occurrence counts of these FGs as handcrafted features and train a specialist decision tree for each property. Note that a decision tree learns from handcrafted features differently and is more sensitive to activity cliffs than LLMs \cite{xia2023understanding}, with its knowledge encoded in the learned predictive rule. We then verbalize this rule and incorporate it into the prompts to distill this knowledge into LLMs. 

To further boost the performance of LLMs, we introduce rule-consistency, a test-time scaling technique inspired by bootstrap aggregating (bagging) \cite{breiman1996bagging}. Particularly, instead of using a single decision tree, we train a Random Forest of $N$ decision trees for each property, resulting in $N$ predictive rules, and use one randomly selected rule per input molecule. The randomness mirrors the inherent diversity in the Random Forest, preventing LLMs from relying on a single rule. During inference, rule-consistency aggregates $N$ predictions generated from $N$ different prompts, each constructed with a different rule, unlike the standard self-consistency \cite{wang2023}, which aggregates $N$ predictions generated from the same prompt. 

To validate the effectiveness of the proposed method, we conduct experiments on the TDC benchmark \cite{huang2021therapeutics} with 22 prediction tasks (corresponding to 22 ADMET properties), including 9 regression tasks and 13 classification tasks. In particular, we use two open-source 2B-parameter LLMs, Gemma-2-2B \cite{team2024gemma} and Granite-3.3-2B \cite{granite2024granite}. The results show that TreeKD enhances the performance of LLMs on most tasks (19/22 for Gemma-2-2B and 19/22 for Granite-3.3-2B), while rule-consistency enhances their performance further (17/22 for Gemma-2-2B and 17/22 for Granite-3.3-2B), significantly surpassing self-consistency. 

We highlight the following key contributions:
\begin{itemize}[topsep=0em, noitemsep, leftmargin=*]
\item \textbf{TreeKD}. We propose TreeKD, a novel knowledge distillation method that incorporates the verbalized knowledge of tree-based specialist models into LLMs. 
\item \textbf{Rule-Consistency}. We additionally introduce rule-consistency, a test-time scaling technique inspired by bagging. 
\item \textbf{Evaluation}. We carry out an extensive evaluation with two LLMs on the TDC benchmark and demonstrate that our method substantially enhances the performance of LLMs. 
\item \textbf{In-Depth Analyses}. Lastly, we conduct comprehensive analyses to understand the impact of various factors on our method’s effectiveness, as well as how LLMs utilize the rule. 
\end{itemize}

\begin{figure*}
\centering
\includegraphics[width=1\linewidth]{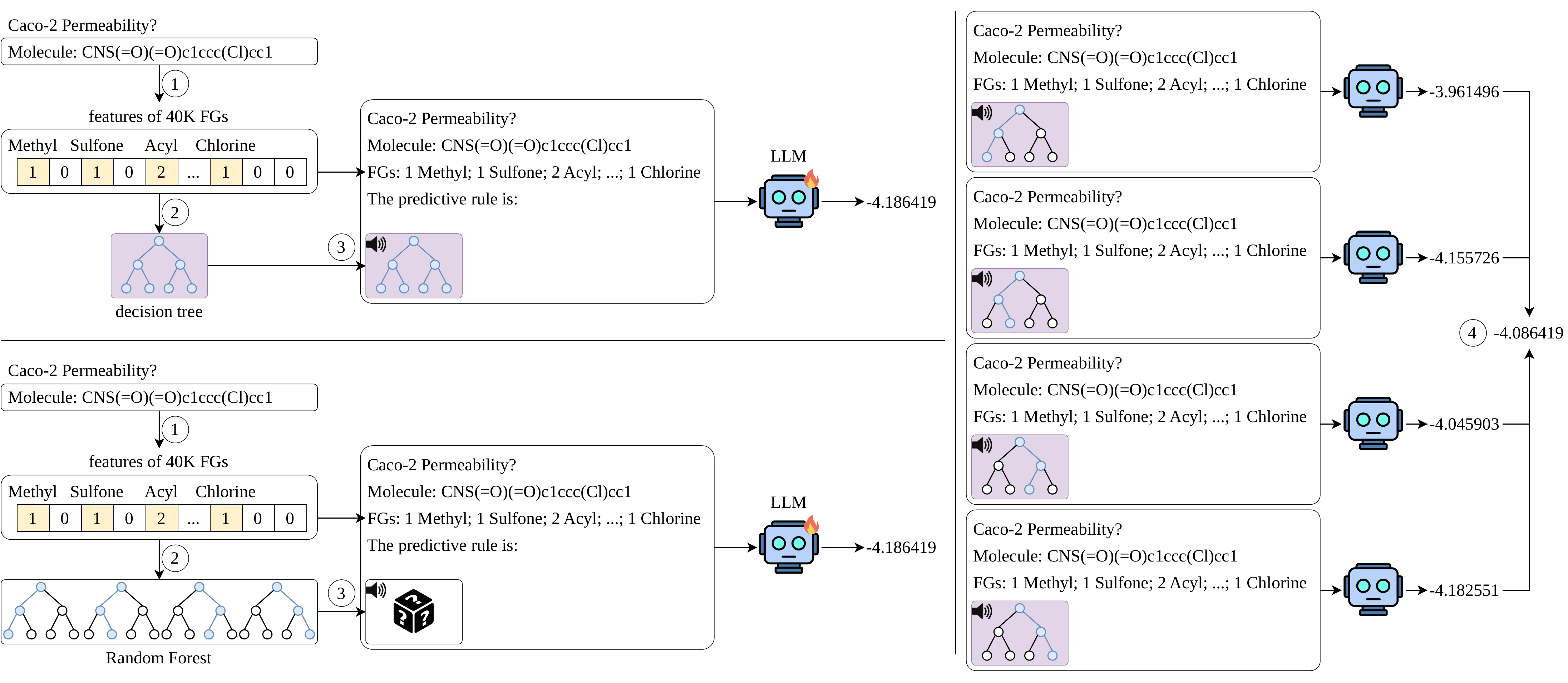}
\caption{
An overview of our proposed method, consisting of 4 stages: \ding{172} Extracting FGs, \ding{173} Building Specialist Models, \ding{174} Distilling Knowledge to LLMs, and \ding{175} Test-Time Scaling. 
}
\label{fig-2}
\end{figure*}

\section{Related Work}

\subsection{Generalist Models for MPP}
Since molecules can be represented using their SMILES string \cite{weininger1988smiles}, LLMs have been applied as generalist models to a variety of drug discovery tasks \cite{lu2025large, zhang2025scientific}, including Molecular Property Prediction (MPP), offering unified and interactive capabilities. However, most existing approaches rely on naive training (solely using the SMILES string in the prompt) \cite{fang2024molinstructions, yu2024llasmol, liu2024moleculargpt, wang2025txgemma}, limiting the full potential of LLMs and causing their performance to significantly lag behind that of strong specialist models. One of the key issues is that LLMs struggle to recognize and utilize important local substructures within the molecules \cite{zhao2023scientific, xia2023understanding, runcie2025assessing}, which largely determine their properties \cite{rogers2010extended, mauri2017molecular}. In light of this, \citet{liu-etal-2023-molca, le2024molx, cao-etal-2025-instructmol} augment LLMs by integrating information from 2D/3D graphs or fingerprints via an external module. Yet, these methods require data- and resource-intensive alignment processes and introduce risks of data leakage. Orthogonally, our method directly harnesses the knowledge of tree-based specialist models on local substructures to assist LLMs and does not require any alignment processes. 

\subsection{Knowledge Distillation with LLMs}
Knowledge Distillation (KD) \cite{hinton2015distilling} is a widely used method for training a neural network with the knowledge of another neural network. Later, \citet{kim2016sequence} introduced SeqKD as a variant of KD tailored for training an LLM with the knowledge of another LLM. Remarkably, Alpaca \cite{alpaca} and Vicuna \cite{vicuna2023} are prominent examples of SeqKD. Nevertheless, none of these methods enable training an LLM with the knowledge of tree-based specialist models, leaving this area largely unexplored \citet{yang2025survey}. 

\subsection{Test-Time Scaling for LLMs}
The purpose of test-time scaling \cite{zhang2025survey, yang2025test} is to boost the performance of an LLM by increasing computation during inference, rather than during training. A classic example of test-time scaling techniques is arguably Chain-of-Thought (CoT) \cite{wei2022chain}, which generates intermediate reasoning steps before generating the prediction. The success of CoT has inspired various extensions such as Tree-of-Thought (ToT) \cite{yao2023} and Graph-of-Thought (GoT) \cite{besta2024graph}. Another line of test-time scaling techniques focuses on generating multiple predictions in parallel and then aggregating them into the final prediction \cite{wang2023, zhu2024path, 2024-self-para}. A notable and simple one is self-consistency \cite{wang2023}, which generates multiple predictions from the same prompt by using a sampling strategy (e.g., temperature sampling or nucleus sampling). Nonetheless, plainly relying on a sampling strategy introduces substantial noise into the predictions, which might hurt performance in certain scenarios. 

\section{Methodology}
This section describes our proposed method in detail, as illustrated in Figure \ref{fig-2}, consisting of four stages: Extracting FGs (Section \ref{sec-3.1}), Building Specialist Models (Section \ref{sec-3.2}), Distilling Knowledge into LLMs (Section \ref{sec-3.3}), and Test-Time Scaling (Section \ref{sec-3.4}). 

First and foremost, let $P$ denote the set of properties under consideration. For each property $p \in P$ (e.g., Caco-2 Permeability), we define the training dataset of $n_p$ samples as $D_p = \{(m_i, y_i)\}_{i=1}^{n_p}$, where $m_i$ represents an input molecule as a SMILES string and $y_i$ denotes its label (i.e., a numerical value for regression tasks, and “Yes” or “No” for classification tasks).

\subsection{Extracting FGs}
\label{sec-3.1}
Given the input molecule $m_i$, we use SMART-Chemist \cite{gutermuth2025smartchemist}, a powerful cheminformatics tool for SMARTS pattern matching \cite{schmidt2019comparing}, to extract the 40K most common functional groups (FGs). This process is deterministic and produces a feature vector $v_i$ that stores the occurrence counts of these FGs in $m_i$.

\subsection{Building Specialist Models}
\label{sec-3.2}
For each property $p \in P$, on the training dataset $D_p^{FGs} = \{(v_i, y_i)\}_{i=1}^{n_p}$, we then train a specialist decision tree $r_p$ (see Figure \ref{fig-2} Top-Left). Unless otherwise specified, the maximum depth ($d$) of the decision tree is set to 6 to balance performance and complexity. Detailed hyperparameters of specialist models are provided in Appendix \ref{app-a}.

\begin{figure}
\centering
\includegraphics[width=1\linewidth]{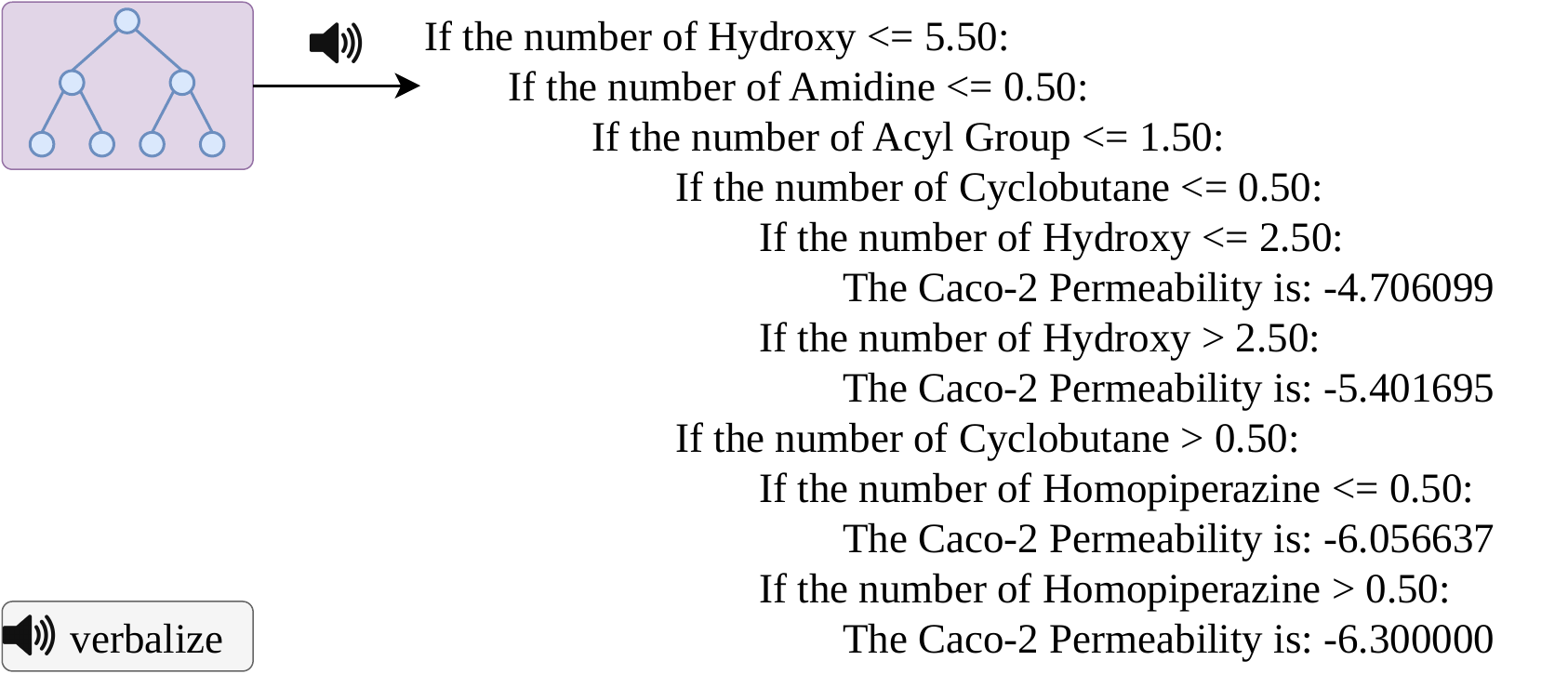}
\caption{
An example of rule verbalization. The decision tree is turned into a hierarchy of if-then conditions, where indentation indicates the depth of nodes. 
}
\label{fig-3}
\end{figure}

\begin{figure}[!t]
\centering
\includegraphics[width=1\linewidth]{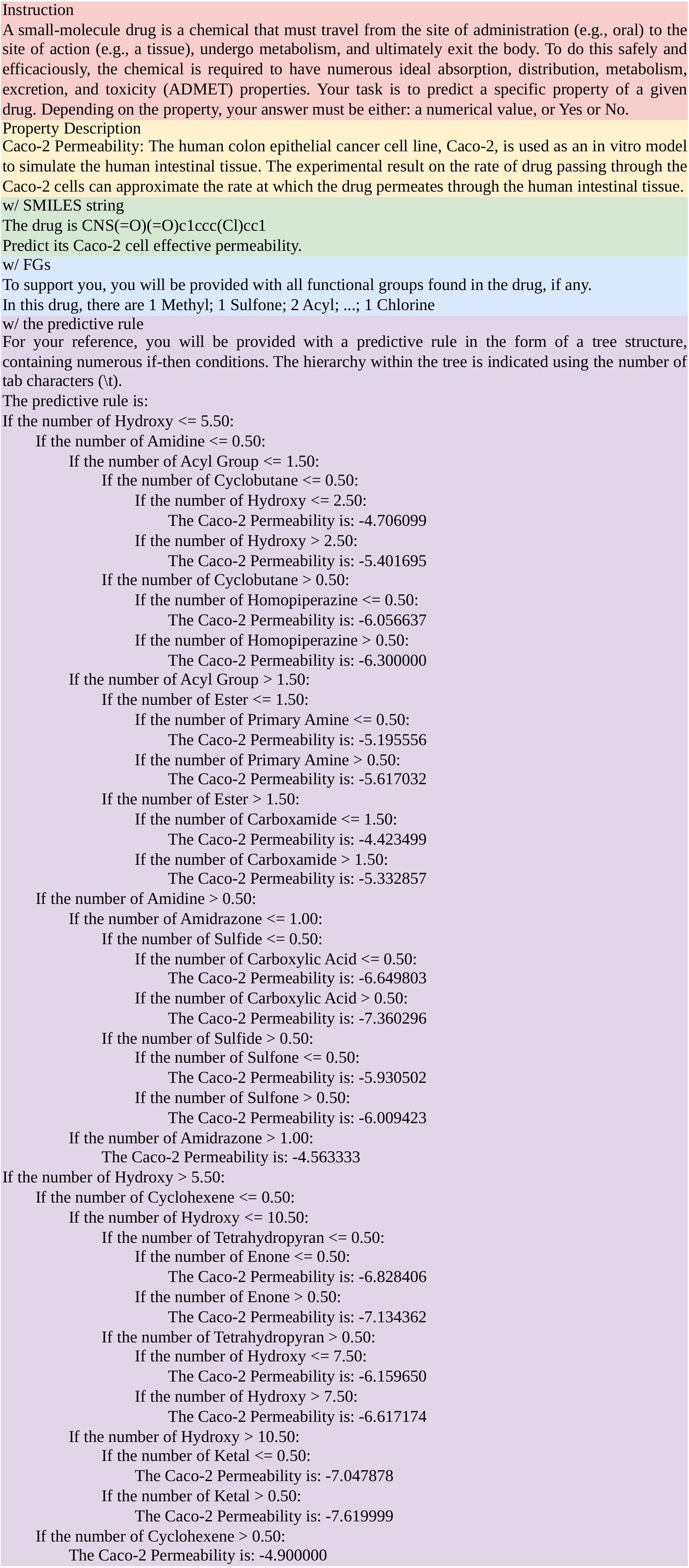}
\caption{
An example of a complete prompt on Caco-2 Permeability. All numerical values are rounded to 6 decimal places. 
}
\label{fig-4}
\end{figure}

\subsection{Distilling Knowledge into LLMs}
\label{sec-3.3}
Unlike conventional KD and its variants, TreeKD uses natural language as a bridge to transfer knowledge from specialist models to LLMs. More specifically, to train an LLM with the knowledge of the decision tree, we exploit the tree's learned predictive rule, which encodes its knowledge of the relationship between FGs and the property. For each property $p \in P$ and the corresponding decision tree $r_p$, we verbalize the tree’s rule by representing it as a hierarchy of if-then conditions that the LLM can digest. The structure of the hierarchy is indicated by the number of tab characters (\texttt{\textbackslash t}) used for indentation. Figure \ref{fig-3} provides an example of how a rule is converted into natural language. We then incorporate the verbalized rule into the prompt associated with the input molecule. 

In particular, for the input molecule $m_i$, the corresponding prompt $q_i$ sequentially includes a general instruction, the description of the property $p$, its SMILES string, its identified FGs, and the verbalized rule. An example of a complete prompt is shown in Figure \ref{fig-4}. Beyond the global information of $m_i$, the prompt $q_i$ provides the LLM with identified FGs as well as the knowledge of the relationship between FGs and the property. The LLM is then trained to use such knowledge. On the joint training dataset $\bigcup_{p \in P} D_p^{LLM}$, with $D_p^{LLM} = \{(q_i, y_i)\}_{i=1}^{n_p}$, we minimize the standard next-token cross-entropy loss:
\begin{equation}
L(\theta) = \sum_{p \in P} \sum_{i=1}^{n_p} \sum_{t=1}^{T_i} - \texttt{log}\text{ }\mathrm{P}_{\theta}( y_i^t\text{ }|\text{ }q_i, y_i^{< t} ), 
\end{equation}
where $\theta$ denotes the LLM's parameters, $T_i$ is the total number of tokens in $y_i$, and $y_i^t$ and $y_i^{< t}$ denote the token at position $t$ and all preceding tokens, respectively. Note that for regression tasks, $y_i$ is a numerical value and is rounded to 6 decimal places to preserve its precision. Detailed hyperparameters for training the LLM are provided in Appendix \ref{app-b}.

\subsection{Test-Time Scaling}
\label{sec-3.4}
Inspired by classical bagging \cite{breiman1996bagging}, we introduce a test-time scaling technique: rule-consistency, which leverages the diversity among decision trees in a Random Forest instead of relying on the sampling strategy used in self-consistency. For each property $p \in P$, we train a Random Forest consisting of $N$ decision trees, yielding $N$ corresponding predictive rules. Unless otherwise specified, we set $N=50$ to ensure sufficient robustness and diversity for the Random Forest. While training the LLM, we randomly select one rule for the input molecule $m_i$ (see Figure \ref{fig-2} Bottom-Left). During inference, $N$ different rules are used to construct $N$ different prompts, denoted as $\{q_{i, k}\}_{k=1}^N$. Then, rule-consistency aggregates $N$ predictions generated from these prompts to return the final prediction (see Figure \ref{fig-2} Right):
\begin{equation}
\widehat{y_i} = \frac{1}{N} \sum_{k=1}^{N} y_{i, k} \text{ } \text{where} \text{ } y_{i, k} \sim \mathrm{P}_{\theta}( y_{i, k}^t\text{ }|\text{ }q_{i, k}, y_{i, k}^{< t} ). 
\end{equation}
For regression tasks, $y_{i, k}$ is a numerical value, while for classification tasks, we use the confidence score corresponding to “Yes” and “No” instead of the label itself. 

\begin{table*}[!t]
\centering\footnotesize\renewcommand{\arraystretch}{1.1079}\setlength\tabcolsep{0pt}
\caption{
The results on the TDC benchmark. (\textcolor{ForestGreen}{green}) and (\textcolor{red}{red}) values for TreeKD indicate performance gaps relative to naive training; \textbf{boldface} denotes better results in a direct comparison with TxGemma2-2B. In the 2 rightmost columns, (\textcolor{ForestGreen}{green}) and (\textcolor{red}{red}) values indicate performance gaps relative to TreeKD for test-time scaling techniques; \underline{underlining} denotes better results between them. The subscripts report standard deviations across 3 runs. 
}
\label{tab-1}
\begin{tabular*}{\linewidth}{@{\extracolsep{\fill}} llr>{\columncolor{gray!20}}c>{\columncolor{gray!20}}cccc@{\hspace{4pt}}|cc}
\midrule
\multirow[t]{2}{*}{Type} &\multirow[t]{2}{*}{Property} &\multirow[t]{2}{*}{Metric} &\multicolumn{2}{>{\columncolor{gray!20}}c}{Specialist Models} &\multirow[t]{2}{*}{TxGemma-2B} &\multicolumn{4}{c}{Gemma-2-2B} \\
\cmidrule{4-5}\cmidrule{7-10}
& & &MapLight &Decision Tree & &Naive &TreeKD &w/ self-consistency &w/ rule-consistency \\
\midrule
\multirow[t]{9}{*}{Reg} 
&Caco-2 Permeability 	&MAE (↓) 	&0.269 &0.413 			&0.629          &0.547 &\textbf{0.476} (\textcolor{ForestGreen}{\texttt{+}0.071}) &0.551\textsubscript{0.008}             (\textcolor{red}{\texttt{-}0.075}) &\underline{0.528}\textsubscript{0.005} (\textcolor{red}{\texttt{-}0.052}) \\
&Lipophilicity 		&MAE (↓) 	&0.547 &0.848 			&0.598          &0.672 &\textbf{0.579} (\textcolor{ForestGreen}{\texttt{+}0.093}) &0.634\textsubscript{0.009}             (\textcolor{red}{\texttt{-}0.055}) &\underline{0.600}\textsubscript{0.006} (\textcolor{red}{\texttt{-}0.021}) \\
&Solubility 		&MAE (↓) 	&0.793 &1.771 			&0.950          &2.921 &\textbf{0.921} (\textcolor{ForestGreen}{\texttt{+}2.000}) &0.982\textsubscript{0.019}             (\textcolor{red}{\texttt{-}0.061}) &\underline{0.868}\textsubscript{0.025} (\textcolor{ForestGreen}{\texttt{+}0.053}) \\
&PPBR 			&MAE (↓) 	&7.654 &11.492\phantom{0} 	&9.986          &7.556 &\textbf{7.503} (\textcolor{ForestGreen}{\texttt{+}0.053}) &8.380\textsubscript{0.168}             (\textcolor{red}{\texttt{-}0.877}) &\underline{7.459}\textsubscript{0.106} (\textcolor{ForestGreen}{\texttt{+}0.044}) \\
&VDss 			&Spearman (↑) 	&0.722 &0.414 			&0.387          &0.522 &\textbf{0.647} (\textcolor{ForestGreen}{\texttt{+}0.125}) &0.602\textsubscript{0.013}             (\textcolor{red}{\texttt{-}0.045}) &\underline{0.675}\textsubscript{0.007} (\textcolor{ForestGreen}{\texttt{+}0.028}) \\
&Half Life 		&Spearman (↑) 	&0.556 &0.007 			&0.022          &0.259 &\textbf{0.272} (\textcolor{ForestGreen}{\texttt{+}0.013}) &\underline{0.385}\textsubscript{0.032} (\textcolor{ForestGreen}{\texttt{+}0.113}) &0.339\textsubscript{0.025}             (\textcolor{ForestGreen}{\texttt{+}0.067}) \\
&Clearance Microsome 	&Spearman (↑) 	&0.615 &0.505 			&0.504          &0.571 &\textbf{0.624} (\textcolor{ForestGreen}{\texttt{+}0.053}) &0.599\textsubscript{0.008}             (\textcolor{red}{\texttt{-}0.025}) &\underline{0.611}\textsubscript{0.010} (\textcolor{red}{\texttt{-}0.013}) \\
&Clearance Hepatocyte 	&Spearman (↑) 	&0.459 &0.244 			&0.346          &0.421 &\textbf{0.419} (\textcolor{red}{\texttt{-}0.002}) &0.360\textsubscript{0.027}             (\textcolor{red}{\texttt{-}0.059}) &\underline{0.412}\textsubscript{0.020} (\textcolor{red}{\texttt{-}0.007}) \\
&LD50 			&MAE (↓) 	&0.620 &0.805 			&0.711          &0.713 &\textbf{0.680} (\textcolor{ForestGreen}{\texttt{+}0.033}) &0.718\textsubscript{0.016}             (\textcolor{red}{\texttt{-}0.038}) &\underline{0.628}\textsubscript{0.009} (\textcolor{ForestGreen}{\texttt{+}0.052}) \\
\midrule
\multirow[t]{13}{*}{Cls} 
&HIA 			&AUROC (↑) 	&0.982 &0.729 			&0.916          &0.820 &\textbf{0.953} (\textcolor{ForestGreen}{\texttt{+}0.133}) &0.956\textsubscript{0.005}             (\textcolor{ForestGreen}{\texttt{+}0.003}) &\underline{0.964}\textsubscript{0.001} (\textcolor{ForestGreen}{\texttt{+}0.011}) \\
&Pgp Inhibition 		&AUROC (↑) 	&0.931 &0.758 			&0.860          &0.818 &\textbf{0.881} (\textcolor{ForestGreen}{\texttt{+}0.063}) &\underline{0.886}\textsubscript{0.005} (\textcolor{ForestGreen}{\texttt{+}0.005}) &0.883\textsubscript{0.002}             (\textcolor{ForestGreen}{\texttt{+}0.002}) \\
&Bioavailability 		&AUROC (↑) 	&0.751 &0.670 			&0.634          &0.692 &\textbf{0.721} (\textcolor{ForestGreen}{\texttt{+}0.029}) &0.725\textsubscript{0.046}             (\textcolor{ForestGreen}{\texttt{+}0.004}) &\underline{0.793}\textsubscript{0.033} (\textcolor{ForestGreen}{\texttt{+}0.072}) \\
&BBB 			&AUROC (↑) 	&0.922 &0.754 			&\textbf{0.869} &0.879 &0.850          (\textcolor{red}{\texttt{-}0.029}) &0.853\textsubscript{0.003}             (\textcolor{ForestGreen}{\texttt{+}0.003}) &\underline{0.906}\textsubscript{0.006} (\textcolor{ForestGreen}{\texttt{+}0.056}) \\
&CYP2D6 Inhibition 	&AUPRC (↑) 	&0.722 &0.381 			&\textbf{0.590} &0.565 &0.573          (\textcolor{ForestGreen}{\texttt{+}0.008}) &0.578\textsubscript{0.006}             (\textcolor{ForestGreen}{\texttt{+}0.005}) &\underline{0.591}\textsubscript{0.001} (\textcolor{ForestGreen}{\texttt{+}0.018}) \\
&CYP3A4 Inhibition 	&AUPRC (↑) 	&0.883 &0.629 			&0.716          &0.778 &\textbf{0.841} (\textcolor{ForestGreen}{\texttt{+}0.063}) &0.847\textsubscript{0.002}             (\textcolor{ForestGreen}{\texttt{+}0.006}) &\underline{0.869}\textsubscript{0.000} (\textcolor{ForestGreen}{\texttt{+}0.028}) \\
&CYP2C9 Inhibition 	&AUPRC (↑) 	&0.785 &0.507 			&0.722          &0.696 &\textbf{0.737} (\textcolor{ForestGreen}{\texttt{+}0.041}) &0.741\textsubscript{0.001}             (\textcolor{ForestGreen}{\texttt{+}0.004}) &\underline{0.763}\textsubscript{0.001} (\textcolor{ForestGreen}{\texttt{+}0.026}) \\
&CYP2D6 Substrate 	&AUPRC (↑) 	&0.713 &0.431 			&0.619          &0.569 &\textbf{0.721} (\textcolor{ForestGreen}{\texttt{+}0.152}) &0.604\textsubscript{0.017}             (\textcolor{red}{\texttt{-}0.117}) &\underline{0.734}\textsubscript{0.024} (\textcolor{ForestGreen}{\texttt{+}0.013}) \\
&CYP3A4 Substrate 	&AUPRC (↑) 	&0.702 &0.612 			&0.685          &0.668 &\textbf{0.740} (\textcolor{ForestGreen}{\texttt{+}0.072}) &0.730\textsubscript{0.019}             (\textcolor{red}{\texttt{-}0.010}) &\underline{0.746}\textsubscript{0.019} (\textcolor{ForestGreen}{\texttt{+}0.006}) \\
&CYP2C9 Substrate 	&AUPRC (↑) 	&0.422 &0.328 			&0.368          &0.553 &\textbf{0.426} (\textcolor{red}{\texttt{-}0.127}) &\underline{0.431}\textsubscript{0.014} (\textcolor{ForestGreen}{\texttt{+}0.005}) &0.428\textsubscript{0.033}             (\textcolor{ForestGreen}{\texttt{+}0.002}) \\
&hERG 			&AUROC (↑) 	&0.883 &0.600 			&\textbf{0.891} &0.749 &0.848          (\textcolor{ForestGreen}{\texttt{+}0.099}) &0.855\textsubscript{0.018}             (\textcolor{ForestGreen}{\texttt{+}0.007}) &\underline{0.866}\textsubscript{0.002} (\textcolor{ForestGreen}{\texttt{+}0.018}) \\
&Ames Mutagenicity 	&AUROC (↑) 	&0.864 &0.680 			&0.792          &0.766 &\textbf{0.819} (\textcolor{ForestGreen}{\texttt{+}0.053}) &0.823\textsubscript{0.015}             (\textcolor{ForestGreen}{\texttt{+}0.004}) &\underline{0.845}\textsubscript{0.002} (\textcolor{ForestGreen}{\texttt{+}0.026}) \\
&DILI 			&AUROC (↑) 	&0.892 &0.641 			&0.764          &0.717 &\textbf{0.828} (\textcolor{ForestGreen}{\texttt{+}0.111}) &\underline{0.831}\textsubscript{0.022} (\textcolor{ForestGreen}{\texttt{+}0.003}) &0.803\textsubscript{0.005}             (\textcolor{red}{\texttt{-}0.025}) \\
\midrule
\midrule
\multirow[t]{2}{*}{Type} &\multirow[t]{2}{*}{Property} &\multirow[t]{2}{*}{Metric} &\multicolumn{2}{>{\columncolor{gray!20}}c}{Specialist Models} &\multirow[t]{2}{*}{TxGemma-2B} &\multicolumn{4}{c}{Granite-3.3-2B} \\
\cmidrule{4-5}\cmidrule{7-10}
& & &MapLight &Decision Tree & &Naive &TreeKD &w/ self-consistency &w/ rule-consistency \\
\midrule
\multirow[t]{9}{*}{Reg} 
&Caco-2 Permeability 	&MAE (↓) 	&0.269 &0.413 			&0.629          &0.497 &\textbf{0.402} (\textcolor{ForestGreen}{\texttt{+}0.095}) &0.470\textsubscript{0.007}             (\textcolor{red}{\texttt{-}0.068}) &\underline{0.432}\textsubscript{0.004} (\textcolor{red}{\texttt{-}0.030}) \\
&Lipophilicity 		&MAE (↓) 	&0.547 &0.848 			&\textbf{0.598} &0.637 &0.620          (\textcolor{ForestGreen}{\texttt{+}0.017}) &0.667\textsubscript{0.010}             (\textcolor{red}{\texttt{-}0.047}) &\underline{0.597}\textsubscript{0.006} (\textcolor{ForestGreen}{\texttt{+}0.023}) \\
&Solubility 		&MAE (↓) 	&0.793 &1.771 			&0.950          &0.913 &\textbf{0.884} (\textcolor{ForestGreen}{\texttt{+}0.029}) &1.025\textsubscript{0.020}             (\textcolor{red}{\texttt{-}0.141}) &\underline{0.855}\textsubscript{0.024} (\textcolor{ForestGreen}{\texttt{+}0.029}) \\
&PPBR 			&MAE (↓) 	&7.654 &11.492\phantom{0} 	&9.986          &8.086 &\textbf{7.259} (\textcolor{ForestGreen}{\texttt{+}0.827}) &8.526\textsubscript{0.171}             (\textcolor{red}{\texttt{-}1.267}) &\underline{8.404}\textsubscript{0.119} (\textcolor{red}{\texttt{-}1.145}) \\
&VDss 			&Spearman (↑) 	&0.722 &0.414 			&0.387          &0.479 &\textbf{0.574} (\textcolor{ForestGreen}{\texttt{+}0.095}) &0.564\textsubscript{0.012}             (\textcolor{red}{\texttt{-}0.010}) &\underline{0.662}\textsubscript{0.007} (\textcolor{ForestGreen}{\texttt{+}0.088}) \\
&Half Life 		&Spearman (↑) 	&0.556 &0.007 			&0.022          &0.293 &\textbf{0.340} (\textcolor{ForestGreen}{\texttt{+}0.047}) &0.324\textsubscript{0.027}             (\textcolor{red}{\texttt{-}0.016}) &\underline{0.362}\textsubscript{0.027} (\textcolor{ForestGreen}{\texttt{+}0.022}) \\
&Clearance Microsome 	&Spearman (↑) 	&0.615 &0.505 			&0.504          &0.528 &\textbf{0.663} (\textcolor{ForestGreen}{\texttt{+}0.135}) &0.592\textsubscript{0.008}             (\textcolor{red}{\texttt{-}0.071}) &\underline{0.611}\textsubscript{0.010} (\textcolor{red}{\texttt{-}0.052}) \\
&Clearance Hepatocyte 	&Spearman (↑) 	&0.459 &0.244 			&0.346          &0.286 &\textbf{0.457} (\textcolor{ForestGreen}{\texttt{+}0.171}) &0.380\textsubscript{0.028}             (\textcolor{red}{\texttt{-}0.077}) &\underline{0.483}\textsubscript{0.023} (\textcolor{ForestGreen}{\texttt{+}0.026}) \\
&LD50 			&MAE (↓) 	&0.620 &0.805 			&0.711          &0.686 &\textbf{0.707} (\textcolor{red}{\texttt{-}0.021}) &0.753\textsubscript{0.017}             (\textcolor{red}{\texttt{-}0.046}) &\underline{0.676}\textsubscript{0.010} (\textcolor{ForestGreen}{\texttt{+}0.031}) \\
\midrule
\multirow[t]{13}{*}{Cls} 
&HIA 			&AUROC (↑) 	&0.982 &0.729 			&0.916          &0.875 &\textbf{0.934} (\textcolor{ForestGreen}{\texttt{+}0.059}) &\underline{0.963}\textsubscript{0.005} (\textcolor{ForestGreen}{\texttt{+}0.029}) &0.957\textsubscript{0.002}             (\textcolor{ForestGreen}{\texttt{+}0.023}) \\
&Pgp Inhibition 		&AUROC (↑) 	&0.931 &0.758 			&\textbf{0.860} &0.834 &0.843          (\textcolor{ForestGreen}{\texttt{+}0.009}) &0.846\textsubscript{0.005}             (\textcolor{ForestGreen}{\texttt{+}0.003}) &\underline{0.900}\textsubscript{0.002} (\textcolor{ForestGreen}{\texttt{+}0.057}) \\
&Bioavailability 		&AUROC (↑) 	&0.751 &0.670 			&0.634          &0.737 &\textbf{0.783} (\textcolor{ForestGreen}{\texttt{+}0.046}) &\underline{0.793}\textsubscript{0.050} (\textcolor{ForestGreen}{\texttt{+}0.010}) &0.788\textsubscript{0.033}             (\textcolor{ForestGreen}{\texttt{+}0.005}) \\
&BBB 			&AUROC (↑) 	&0.922 &0.754 			&0.869          &0.833 &\textbf{0.870} (\textcolor{ForestGreen}{\texttt{+}0.037}) &0.864\textsubscript{0.003}             (\textcolor{red}{\texttt{-}0.006}) &\underline{0.878}\textsubscript{0.006} (\textcolor{ForestGreen}{\texttt{+}0.008}) \\
&CYP2D6 Inhibition 	&AUPRC (↑) 	&0.722 &0.381 			&0.590          &0.623 &\textbf{0.600} (\textcolor{red}{\texttt{-}0.023}) &0.605\textsubscript{0.006}             (\textcolor{ForestGreen}{\texttt{+}0.005}) &\underline{0.610}\textsubscript{0.001} (\textcolor{ForestGreen}{\texttt{+}0.010}) \\
&CYP3A4 Inhibition 	&AUPRC (↑) 	&0.883 &0.629 			&0.716          &0.810 &\textbf{0.835} (\textcolor{ForestGreen}{\texttt{+}0.025}) &0.842\textsubscript{0.002}             (\textcolor{ForestGreen}{\texttt{+}0.007}) &\underline{0.849}\textsubscript{0.001} (\textcolor{ForestGreen}{\texttt{+}0.014}) \\
&CYP2C9 Inhibition 	&AUPRC (↑) 	&0.785 &0.507 			&0.722          &0.731 &\textbf{0.750} (\textcolor{ForestGreen}{\texttt{+}0.019}) &0.756\textsubscript{0.001}             (\textcolor{ForestGreen}{\texttt{+}0.006}) &\underline{0.772}\textsubscript{0.001} (\textcolor{ForestGreen}{\texttt{+}0.022}) \\
&CYP2D6 Substrate 	&AUPRC (↑) 	&0.713 &0.431 			&\textbf{0.619} &0.488 &0.589          (\textcolor{ForestGreen}{\texttt{+}0.101}) &0.593\textsubscript{0.017}             (\textcolor{ForestGreen}{\texttt{+}0.004}) &\underline{0.609}\textsubscript{0.020} (\textcolor{ForestGreen}{\texttt{+}0.020}) \\
&CYP3A4 Substrate 	&AUPRC (↑) 	&0.702 &0.612 			&0.685          &0.652 &\textbf{0.753} (\textcolor{ForestGreen}{\texttt{+}0.101}) &\underline{0.757}\textsubscript{0.020} (\textcolor{ForestGreen}{\texttt{+}0.004}) &0.719\textsubscript{0.018}             (\textcolor{red}{\texttt{-}0.034}) \\
&CYP2C9 Substrate 	&AUPRC (↑) 	&0.422 &0.328 			&0.368          &0.461 &\textbf{0.398} (\textcolor{red}{\texttt{-}0.063}) &0.405\textsubscript{0.013}             (\textcolor{ForestGreen}{\texttt{+}0.007}) &\underline{0.487}\textsubscript{0.037} (\textcolor{ForestGreen}{\texttt{+}0.089}) \\
&hERG 			&AUROC (↑) 	&0.883 &0.600 			&\textbf{0.891} &0.720 &0.780          (\textcolor{ForestGreen}{\texttt{+}0.060}) &\underline{0.793}\textsubscript{0.017} (\textcolor{ForestGreen}{\texttt{+}0.013}) &0.783\textsubscript{0.003}             (\textcolor{ForestGreen}{\texttt{+}0.003}) \\
&Ames Mutagenicity 	&AUROC (↑) 	&0.864 &0.680 			&0.792          &0.767 &\textbf{0.826} (\textcolor{ForestGreen}{\texttt{+}0.059}) &0.830\textsubscript{0.015}             (\textcolor{ForestGreen}{\texttt{+}0.004}) &\underline{0.831}\textsubscript{0.003} (\textcolor{ForestGreen}{\texttt{+}0.005}) \\
&DILI 			&AUROC (↑) 	&0.892 &0.641 			&0.764          &0.752 &\textbf{0.817} (\textcolor{ForestGreen}{\texttt{+}0.065}) &\underline{0.821}\textsubscript{0.021} (\textcolor{ForestGreen}{\texttt{+}0.004}) &0.798\textsubscript{0.005}             (\textcolor{red}{\texttt{-}0.019}) \\
\midrule
\end{tabular*}
\end{table*}
\section{Evaluation}

\textbf{Setup}. To validate the effectiveness of the proposed method, we use two open-source LLMs in the 2B-parameter class, Gemma-2-2B \cite{team2024gemma} and Granite-3.3-2B \cite{granite2024granite}, and conduct experiments on the TDC benchmark \cite{huang2021therapeutics}, which consists of 22 prediction tasks, including 9 regression tasks (Reg) and 13 classification tasks (Cls). The corresponding 22 ADMET properties are listed in Appendix \ref{app-c}. The dataset for each task is split into training, validation, and test sets (see Table \ref{tab-4} for details). We use training sets for training LLMs, validation sets for tuning hyperparameters, and report the results on test sets. 

\textbf{Results}. Table \ref{tab-1} presents the results using Gemma-2-2B and Granite-3.3-2B. We draw the following observations:
\begin{itemize}[topsep=0em, noitemsep, leftmargin=*]
\item Compared to naive training (solely using the SMILES string in the prompt), TreeKD enhances LLMs’ performance on most tasks (19/22 for Gemma-2-2B and 19/22 for Granite-3.3-2B), as indicated in (\textcolor{ForestGreen}{green}). For instance, performance gains on PPBR are +0.053 and +0.827. Notably, despite the decision tree’s limited performance due to limited configurations, distilling its knowledge into LLMs still yields clear improvements. While the decision tree achieves only 11.492 on PPBR, TreeKD enables Gemma-2-2B and Granite-3.3-2B to achieve 7.503 and 7.259. This finding supports our hypothesis that the knowledge of the decision tree complements the internal knowledge of LLMs. 
\item In a head-to-head comparison with TxGemma-2-2B \cite{wang2025txgemma}, the strongest generalist model for MPP in the 2B-parameter class (built on top of Gemma-2-2B), TreeKD enables LLMs to achieve superior performance on most tasks (19/22 for Gemma-2-2B and 18/22 for Granite-3.3-2B), as highlighted in \textbf{bold}. Notably, TreeKD significantly narrows the gap with MapLight models \cite{notwell2023admet}, state-of-the-art specialist models on the TDC benchmark. For instance, on PPBR, the gap markedly decreases from -2.364 (for TxGemma-2-2B) to -0.151 and -0.395 for Gemma-2-2B and Granite-3.3-2B. 
\item On top of TreeKD, self-consistency hurts the performance of LLMs on most regression tasks and only enhances their performance on a few classification tasks. In contrast, rule-consistency further enhances the performance on both regression tasks and classification tasks. Overall, rule-consistency clearly outperforms self-consistency on most tasks (18/22 for Gemma-2-2B and 17/22 for Granite-3.3-2B), as \underline{underlined}. 
\end{itemize}

\section{Ablation Studies}

\begin{table}[!t]
\centering\footnotesize\renewcommand{\arraystretch}{1.1079}\setlength\tabcolsep{0pt}
\caption{
The performance gains on the TDC benchmark with evolved prompt variants. (\textcolor{ForestGreen}{green}) and (\textcolor{red}{red}) values indicate performance gaps relative to the preceding one. 
}
\label{tab-2}
\begin{tabular*}{\linewidth}{@{\extracolsep{\fill}} llccc }
\midrule
\multirow[t]{4}{*}{Type} & \multirow[t]{4}{*}{Property} & \multicolumn{3}{c}{Gemma-2-2B} \\\cmidrule{3-5}
\multicolumn{2}{l}{}			& \multicolumn{1}{l}{\cellcolor{myG}Naive} & \multicolumn{1}{l}{\cellcolor{myG}Naive}  & \multicolumn{1}{l}{\cellcolor{myG}Naive}		\\
\multicolumn{2}{l}{}			& \multicolumn{1}{l}{}      		& \multicolumn{1}{l}{\cellcolor{myB}w/ FGs} & \multicolumn{1}{l}{\cellcolor{myB}w/ FGs} 		\\
\multicolumn{2}{l}{}			& \multicolumn{1}{l}{}      		& \multicolumn{1}{l}{}       		 & \multicolumn{1}{l}{\cellcolor{myP}w/ a predictive rule}	\\
\midrule
\multirow[t]{9}{*}{Reg}
& Caco-2 Permeability &0.547 &\textcolor{ForestGreen}{\texttt{+}0.055} &\textcolor{ForestGreen}{\texttt{+}0.016} \\
& Lipophilicity &0.672 &\textcolor{red}{\texttt{-}0.005} &\textcolor{ForestGreen}{\texttt{+}0.098} \\
& Solubility &2.921 &\textcolor{ForestGreen}{\texttt{+}2.012} &\textcolor{red}{\texttt{-}0.012} \\
& PPBR &7.556 &\textcolor{red}{\texttt{-}1.657} &\textcolor{ForestGreen}{\texttt{+}1.710} \\
& VDss &0.522 &\textcolor{ForestGreen}{\texttt{+}0.043} &\textcolor{ForestGreen}{\texttt{+}0.082} \\
& Half Life &0.259 &\textcolor{ForestGreen}{\texttt{+}0.158} &\textcolor{red}{\texttt{-}0.145} \\
& Clearance Microsome &0.571 &\textcolor{ForestGreen}{\texttt{+}0.026} &\textcolor{ForestGreen}{\texttt{+}0.027} \\
& Clearance Hepatocyte &0.421 &\textcolor{red}{\texttt{-}0.135} &\textcolor{ForestGreen}{\texttt{+}0.133} \\
& LD50 &0.713 &\textcolor{red}{\texttt{-}0.007} &\textcolor{ForestGreen}{\texttt{+}0.040} \\
\midrule
\multirow[t]{13}{*}{Cls}
& HIA &0.820 &\textcolor{ForestGreen}{\texttt{+}0.065} &\textcolor{ForestGreen}{\texttt{+}0.068} \\
& Pgp Inhibition &0.818 &\textcolor{ForestGreen}{\texttt{+}0.049} &\textcolor{ForestGreen}{\texttt{+}0.014} \\
& Bioavailability &0.692 &\textcolor{ForestGreen}{\texttt{+}0.059} &\textcolor{red}{\texttt{-}0.030} \\
& BBB &0.879 &\textcolor{red}{\texttt{-}0.026} &\textcolor{red}{\texttt{-}0.003} \\
& CYP2D6 Inhibition &0.565 &\textcolor{ForestGreen}{\texttt{+}0.028} &\textcolor{red}{\texttt{-}0.020} \\
& CYP3A4 Inhibition &0.778 &\textcolor{ForestGreen}{\texttt{+}0.040} &\textcolor{ForestGreen}{\texttt{+}0.023} \\
& CYP2C9 Inhibition &0.696 &\textcolor{ForestGreen}{\texttt{+}0.023} &\textcolor{ForestGreen}{\texttt{+}0.018} \\
& CYP2D6 Substrate &0.569 &\textcolor{ForestGreen}{\texttt{+}0.023} &\textcolor{ForestGreen}{\texttt{+}0.129} \\
& CYP3A4 Substrate &0.668 &\textcolor{ForestGreen}{\texttt{+}0.046} &\textcolor{ForestGreen}{\texttt{+}0.026} \\
& CYP2C9 Substrate &0.553 &\textcolor{red}{\texttt{-}0.109} &\textcolor{red}{\texttt{-}0.018} \\
& hERG &0.749 &\textcolor{ForestGreen}{\texttt{+}0.041} &\textcolor{ForestGreen}{\texttt{+}0.058} \\
& Ames Mutagenicity &0.766 &\textcolor{ForestGreen}{\texttt{+}0.029} &\textcolor{ForestGreen}{\texttt{+}0.024} \\
& DILI &0.717 &\textcolor{ForestGreen}{\texttt{+}0.051} &\textcolor{ForestGreen}{\texttt{+}0.060} \\
\midrule
\midrule
\multirow[t]{4}{*}{Type} & \multirow[t]{4}{*}{Property} & \multicolumn{3}{c}{Granite-3.3-2B} \\\cmidrule{3-5}
\multicolumn{2}{l}{}			& \multicolumn{1}{l}{\cellcolor{myG}Naive} & \multicolumn{1}{l}{\cellcolor{myG}Naive}  & \multicolumn{1}{l}{\cellcolor{myG}Naive}		\\
\multicolumn{2}{l}{}			& \multicolumn{1}{l}{}      		& \multicolumn{1}{l}{\cellcolor{myB}w/ FGs} & \multicolumn{1}{l}{\cellcolor{myB}w/ FGs} 		\\
\multicolumn{2}{l}{}			& \multicolumn{1}{l}{}      		& \multicolumn{1}{l}{}       		 & \multicolumn{1}{l}{\cellcolor{myP}w/ a predictive rule}	\\
\midrule
\multirow[t]{9}{*}{Reg}
& Caco-2 Permeability &0.497 &\textcolor{ForestGreen}{\texttt{+}0.072} &\textcolor{ForestGreen}{\texttt{+}0.023} \\
& Lipophilicity &0.637 &\textcolor{ForestGreen}{\texttt{+}0.004} &\textcolor{ForestGreen}{\texttt{+}0.013} \\
& Solubility &0.913 &\textcolor{ForestGreen}{\texttt{+}0.006} &\textcolor{ForestGreen}{\texttt{+}0.023} \\
& PPBR &8.086 &\textcolor{red}{\texttt{-}1.024} &\textcolor{ForestGreen}{\texttt{+}1.851} \\
& VDss &0.479 &\textcolor{ForestGreen}{\texttt{+}0.059} &\textcolor{ForestGreen}{\texttt{+}0.036} \\
& Half Life &0.293 &\textcolor{red}{\texttt{-}0.020} &\textcolor{ForestGreen}{\texttt{+}0.067} \\
& Clearance Microsome &0.528 &\textcolor{ForestGreen}{\texttt{+}0.078} &\textcolor{ForestGreen}{\texttt{+}0.057} \\
& Clearance Hepatocyte &0.286 &\textcolor{ForestGreen}{\texttt{+}0.017} &\textcolor{ForestGreen}{\texttt{+}0.154} \\
& LD50 &0.686 &\textcolor{red}{\texttt{-}0.029} &\textcolor{ForestGreen}{\texttt{+}0.008} \\
\midrule
\multirow[t]{13}{*}{Cls}
& HIA &0.875 &\textcolor{ForestGreen}{\texttt{+}0.064} &\textcolor{red}{\texttt{-}0.005} \\
& Pgp Inhibition &0.834 &\textcolor{ForestGreen}{\texttt{+}0.012} &\textcolor{red}{\texttt{-}0.003} \\
& Bioavailability &0.737 &\textcolor{ForestGreen}{\texttt{+}0.005} &\textcolor{ForestGreen}{\texttt{+}0.041} \\
& BBB &0.833 &\textcolor{red}{\texttt{-}0.024} &\textcolor{ForestGreen}{\texttt{+}0.061} \\
& CYP2D6 Inhibition &0.623 &\textcolor{red}{\texttt{-}0.025} &\textcolor{ForestGreen}{\texttt{+}0.002} \\
& CYP3A4 Inhibition &0.810 &\textcolor{ForestGreen}{\texttt{+}0.007} &\textcolor{ForestGreen}{\texttt{+}0.018} \\
& CYP2C9 Inhibition &0.731 &\textcolor{ForestGreen}{\texttt{+}0.016} &\textcolor{ForestGreen}{\texttt{+}0.003} \\
& CYP2D6 Substrate &0.488 &\textcolor{ForestGreen}{\texttt{+}0.068} &\textcolor{ForestGreen}{\texttt{+}0.033} \\
& CYP3A4 Substrate &0.652 &\textcolor{ForestGreen}{\texttt{+}0.002} &\textcolor{ForestGreen}{\texttt{+}0.099} \\
& CYP2C9 Substrate &0.461 &\textcolor{red}{\texttt{-}0.093} &\textcolor{ForestGreen}{\texttt{+}0.030} \\
& hERG &0.720 &\textcolor{ForestGreen}{\texttt{+}0.091} &\textcolor{red}{\texttt{-}0.031} \\
& Ames Mutagenicity &0.767 &\textcolor{ForestGreen}{\texttt{+}0.037} &\textcolor{ForestGreen}{\texttt{+}0.022} \\
& DILI &0.752 &\textcolor{ForestGreen}{\texttt{+}0.033} &\textcolor{ForestGreen}{\texttt{+}0.032} \\
\midrule
\end{tabular*}\end{table}
\subsection{Impact of components in the prompt}
\textbf{Setup}. To understand the impact of each component in the prompt for an input molecule, including the SMILES string, the identified FGs, and the predictive rule, we conduct ablation experiments using progressively evolved prompt variants and sequentially measure performance gains. Note that solely using the SMILES string in the prompt reflects naive training. 

\textbf{Results}. Table \ref{tab-2} presents such performance gains. Notably, providing LLMs with all FGs found in the input molecule enhances their performance on most tasks (16/22 for Gemma-2-2B and 16/22 for Granite-3.3-2B). This is likely because LLMs activate their pretrained understanding of FGs when these are presented via standardized names. On top of that, providing LLMs with the predictive rule (TreeKD) also enhances the performance of LLMs on most tasks (16/22 for Gemma-2-2B and 19/22 for Granite-3.3-2B). For instance, performance gains of TreeKD over naive training on Ames Mutagenicity are +0.053 and +0.059, of which +0.029 and +0.037, respectively, are attributable to providing LLMs with all FGs, while +0.024 and +0.022 come from the predictive rule. However, we also notice consistent harms on certain tasks such as LD50 and BBB, where the properties lean more toward global information than FGs \cite{gadaleta2019sar, grant2025blood}. 

\begin{table}[!t]
\centering\footnotesize\renewcommand{\arraystretch}{1.1079}\setlength\tabcolsep{0pt}
\caption{
The results of TreeKD on the TDC benchmark using maximum depths ($d$) of 3 and 6. (\textcolor{ForestGreen}{green}) and (\textcolor{red}{red}) values indicate performance gaps when using $d=3$ relative to using $d=6$. 
}
\label{tab-3}
\begin{tabular*}{\linewidth}{@{\extracolsep{\fill}} llcccc }
\midrule
\multirow[t]{3}{*}{Type} & \multirow[t]{3}{*}{Property} & \multicolumn{2}{c}{Specialist Models} & \multicolumn{2}{c}{Gemma-2-2B} \\\cmidrule{3-4}\cmidrule{5-6}
\multicolumn{2}{l}{} & \multicolumn{2}{c}{Decision Tree} & \multicolumn{2}{c}{TreeKD} \\\cmidrule{3-4}\cmidrule{5-6}
\multicolumn{2}{l}{} & $d$ = 3 & $d$ = 6 & $d$ = 3 & $d$ = 6 \\
\midrule
\multirow[t]{9}{*}{Reg}
&Caco-2 Permeability  &0.566             &\phantom{0}0.413             (\textcolor{ForestGreen}{\texttt{+}0.153}) &0.442 &0.476 (\textcolor{red}{\texttt{-}0.034})         \\
&Lipophilicity        &0.913             &\phantom{0}0.848             (\textcolor{ForestGreen}{\texttt{+}0.065}) &0.636 &0.579 (\textcolor{ForestGreen}{\texttt{+}0.057}) \\
&Solubility           &1.863             &\phantom{0}1.771             (\textcolor{ForestGreen}{\texttt{+}0.092}) &0.949 &0.921 (\textcolor{ForestGreen}{\texttt{+}0.028}) \\
&PPBR                 &11.734\phantom{0} &11.492                       (\textcolor{ForestGreen}{\texttt{+}0.242}) &7.827 &7.503 (\textcolor{ForestGreen}{\texttt{+}0.324}) \\
&VDss                 &0.040             &\phantom{0}0.414             (\textcolor{ForestGreen}{\texttt{+}0.374}) &0.570 &0.647 (\textcolor{ForestGreen}{\texttt{+}0.077}) \\
&Half Life            &0.137             &\phantom{0}0.007             (\textcolor{red}{\texttt{-}0.130})         &0.386 &0.272 (\textcolor{red}{\texttt{-}0.114})         \\
&Clearance Microsome  &0.500             &\phantom{0}0.505             (\textcolor{ForestGreen}{\texttt{+}0.005}) &0.566 &0.624 (\textcolor{ForestGreen}{\texttt{+}0.058}) \\
&Clearance Hepatocyte &0.231             &\phantom{0}0.244             (\textcolor{ForestGreen}{\texttt{+}0.013}) &0.256 &0.419 (\textcolor{ForestGreen}{\texttt{+}0.163}) \\
&LD50                 &0.819             &\phantom{0}0.805             (\textcolor{ForestGreen}{\texttt{+}0.014}) &0.689 &0.680 (\textcolor{ForestGreen}{\texttt{+}0.009}) \\
\midrule
\multirow[t]{13}{*}{Cls}
&HIA                  &0.827             &\phantom{0}0.729             (\textcolor{red}{\texttt{-}0.098})         &0.919 &0.953 (\textcolor{ForestGreen}{\texttt{+}0.034}) \\
&Pgp Inhibition       &0.748             &\phantom{0}0.758             (\textcolor{ForestGreen}{\texttt{+}0.010}) &0.901 &0.881 (\textcolor{red}{\texttt{-}0.020})         \\
&Bioavailability      &0.592             &\phantom{0}0.670             (\textcolor{ForestGreen}{\texttt{+}0.078}) &0.710 &0.721 (\textcolor{ForestGreen}{\texttt{+}0.011}) \\
&BBB                  &0.732             &\phantom{0}0.754             (\textcolor{ForestGreen}{\texttt{+}0.022}) &0.817 &0.850 (\textcolor{ForestGreen}{\texttt{+}0.033}) \\
&CYP2D6 Inhibition    &0.369             &\phantom{0}0.381             (\textcolor{ForestGreen}{\texttt{+}0.012}) &0.534 &0.573 (\textcolor{ForestGreen}{\texttt{+}0.039}) \\
&CYP3A4 Inhibition    &0.538             &\phantom{0}0.629             (\textcolor{ForestGreen}{\texttt{+}0.091}) &0.808 &0.841 (\textcolor{ForestGreen}{\texttt{+}0.033}) \\
&CYP2C9 Inhibition    &0.455             &\phantom{0}0.507             (\textcolor{ForestGreen}{\texttt{+}0.052}) &0.713 &0.737 (\textcolor{ForestGreen}{\texttt{+}0.024}) \\
&CYP2D6 Substrate     &0.445             &\phantom{0}0.431             (\textcolor{red}{\texttt{-}0.014})         &0.608 &0.721 (\textcolor{ForestGreen}{\texttt{+}0.113}) \\
&CYP3A4 Substrate     &0.584             &\phantom{0}0.612             (\textcolor{ForestGreen}{\texttt{+}0.028}) &0.733 &0.740 (\textcolor{ForestGreen}{\texttt{+}0.007}) \\
&CYP2C9 Substrate     &0.293             &\phantom{0}0.328             (\textcolor{ForestGreen}{\texttt{+}0.035}) &0.313 &0.426 (\textcolor{ForestGreen}{\texttt{+}0.113}) \\
&hERG                 &0.616             &\phantom{0}0.600             (\textcolor{red}{\texttt{-}0.016})         &0.807 &0.848 (\textcolor{ForestGreen}{\texttt{+}0.041}) \\
&Ames Mutagenicity    &0.617             &\phantom{0}0.680             (\textcolor{ForestGreen}{\texttt{+}0.063}) &0.798 &0.819 (\textcolor{ForestGreen}{\texttt{+}0.021}) \\
&DILI                 &0.604             &\phantom{0}0.641             (\textcolor{ForestGreen}{\texttt{+}0.037}) &0.756 &0.828 (\textcolor{ForestGreen}{\texttt{+}0.072}) \\
\midrule
\midrule
\multirow[t]{3}{*}{Type} & \multirow[t]{3}{*}{Property} & \multicolumn{2}{c}{Specialist Models} & \multicolumn{2}{c}{Granite-3.3-2B} \\\cmidrule{3-4}\cmidrule{5-6}
\multicolumn{2}{l}{} & \multicolumn{2}{c}{Decision Tree} & \multicolumn{2}{c}{TreeKD} \\\cmidrule{3-4}\cmidrule{5-6}
\multicolumn{2}{l}{} & $d$ = 3 & $d$ = 6 & $d$ = 3 & $d$ = 6 \\
\midrule
\multirow[t]{9}{*}{Reg}
&Caco-2 Permeability  &0.566             &\phantom{0}0.413             (\textcolor{ForestGreen}{\texttt{+}0.153}) &0.501 &0.402 (\textcolor{ForestGreen}{\texttt{+}0.099}) \\
&Lipophilicity        &0.913             &\phantom{0}0.848             (\textcolor{ForestGreen}{\texttt{+}0.065}) &0.666 &0.620 (\textcolor{ForestGreen}{\texttt{+}0.046}) \\
&Solubility           &1.863             &\phantom{0}1.771             (\textcolor{ForestGreen}{\texttt{+}0.092}) &0.917 &0.884 (\textcolor{ForestGreen}{\texttt{+}0.033}) \\
&PPBR                 &11.734\phantom{0} &11.492                       (\textcolor{ForestGreen}{\texttt{+}0.242}) &9.226 &7.259 (\textcolor{ForestGreen}{\texttt{+}1.967}) \\
&VDss                 &0.040             &\phantom{0}0.414             (\textcolor{ForestGreen}{\texttt{+}0.374}) &0.521 &0.574 (\textcolor{ForestGreen}{\texttt{+}0.053}) \\
&Half Life            &0.137             &\phantom{0}0.007             (\textcolor{red}{\texttt{-}0.130})         &0.388 &0.340 (\textcolor{red}{\texttt{-}0.048})         \\
&Clearance Microsome  &0.500             &\phantom{0}0.505             (\textcolor{ForestGreen}{\texttt{+}0.005}) &0.390 &0.663 (\textcolor{ForestGreen}{\texttt{+}0.273}) \\
&Clearance Hepatocyte &0.231             &\phantom{0}0.244             (\textcolor{ForestGreen}{\texttt{+}0.013}) &0.206 &0.457 (\textcolor{ForestGreen}{\texttt{+}0.251}) \\
&LD50                 &0.819             &\phantom{0}0.805             (\textcolor{ForestGreen}{\texttt{+}0.014}) &0.745 &0.707 (\textcolor{ForestGreen}{\texttt{+}0.038}) \\
\midrule
\multirow[t]{13}{*}{Cls}
&HIA                  &0.827             &\phantom{0}0.729             (\textcolor{red}{\texttt{-}0.098})         &0.772 &0.934 (\textcolor{ForestGreen}{\texttt{+}0.162}) \\
&Pgp Inhibition       &0.748             &\phantom{0}0.758             (\textcolor{ForestGreen}{\texttt{+}0.010}) &0.850 &0.843 (\textcolor{red}{\texttt{-}0.007})         \\
&Bioavailability      &0.592             &\phantom{0}0.670             (\textcolor{ForestGreen}{\texttt{+}0.078}) &0.764 &0.783 (\textcolor{ForestGreen}{\texttt{+}0.019}) \\
&BBB                  &0.732             &\phantom{0}0.754             (\textcolor{ForestGreen}{\texttt{+}0.022}) &0.861 &0.870 (\textcolor{ForestGreen}{\texttt{+}0.009}) \\
&CYP2D6 Inhibition    &0.369             &\phantom{0}0.381             (\textcolor{ForestGreen}{\texttt{+}0.012}) &0.581 &0.600 (\textcolor{ForestGreen}{\texttt{+}0.019}) \\
&CYP3A4 Inhibition    &0.538             &\phantom{0}0.629             (\textcolor{ForestGreen}{\texttt{+}0.091}) &0.808 &0.835 (\textcolor{ForestGreen}{\texttt{+}0.027}) \\
&CYP2C9 Inhibition    &0.455             &\phantom{0}0.507             (\textcolor{ForestGreen}{\texttt{+}0.052}) &0.738 &0.750 (\textcolor{ForestGreen}{\texttt{+}0.012}) \\
&CYP2D6 Substrate     &0.445             &\phantom{0}0.431             (\textcolor{red}{\texttt{-}0.014})         &0.532 &0.589 (\textcolor{ForestGreen}{\texttt{+}0.057}) \\
&CYP3A4 Substrate     &0.584             &\phantom{0}0.612             (\textcolor{ForestGreen}{\texttt{+}0.028}) &0.683 &0.753 (\textcolor{ForestGreen}{\texttt{+}0.070}) \\
&CYP2C9 Substrate     &0.293             &\phantom{0}0.328             (\textcolor{ForestGreen}{\texttt{+}0.035}) &0.387 &0.398 (\textcolor{ForestGreen}{\texttt{+}0.011}) \\
&hERG                 &0.616             &\phantom{0}0.600             (\textcolor{red}{\texttt{-}0.016})         &0.772 &0.780 (\textcolor{ForestGreen}{\texttt{+}0.008}) \\
&Ames Mutagenicity    &0.617             &\phantom{0}0.680             (\textcolor{ForestGreen}{\texttt{+}0.063}) &0.802 &0.826 (\textcolor{ForestGreen}{\texttt{+}0.024}) \\
&DILI                 &0.604             &\phantom{0}0.641             (\textcolor{ForestGreen}{\texttt{+}0.037}) &0.828 &0.817 (\textcolor{red}{\texttt{-}0.011})         \\
\midrule
\end{tabular*}\end{table}
\subsection{Impact of the decision tree’s complexity}
\textbf{Setup}. We also investigate the impact of the decision tree’s complexity on TreeKD’s effectiveness by reducing its maximum depth ($d$) from 6 to 3. 

\textbf{Results}. Table \ref{tab-3} presents such results. Overall, increasing the maximum depth ($d$) generally produces a better decision tree (on 18/22 tasks). More importantly, this improvement typically translates into better performance of LLMs on most tasks (19/22 for Gemma-2-2B and 19/22 for Granite-3.3-2B), suggesting a positive impact of the decision tree’s complexity on TreeKD’s effectiveness. For instance, on Ames Mutagenicity, an improvement of +0.063 in the decision tree results in improvements of +0.021 and +0.024 in Gemma-2-2B and Granite-3.3-2B, respectively. However, we also observe a few exceptions where a better decision tree unexpectedly hurts LLMs’ performance (on Pgp Inhibition) or where LLMs’ performance is still improved regardless of a weaker decision tree (on HIA and hERG).

\section{Analysis of Rule Utilization}

\begin{figure}
\centering
\includegraphics[width=1\linewidth]{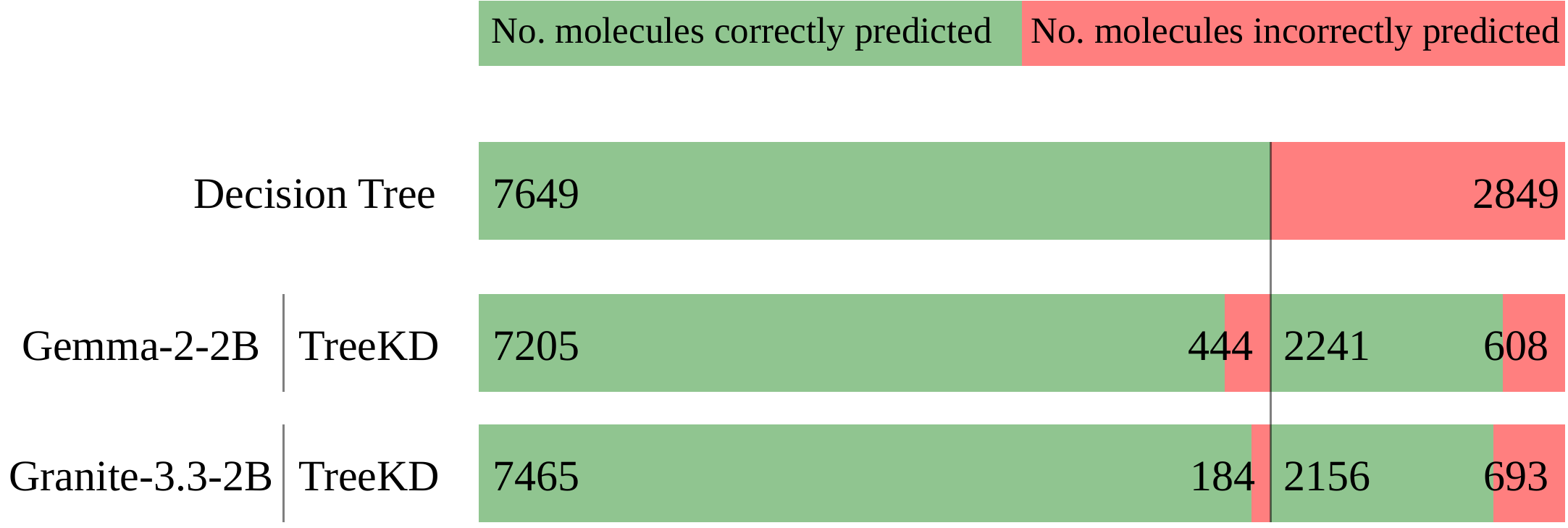}
\caption{
The distribution of 10498 molecules in the test set of 13 classification tasks from the TDC benchmark, according to whether they are correctly predicted by the decision tree and the LLMs. 
}
\label{fig-5}
\end{figure}

Although the knowledge of the decision tree complements the internal knowledge of LLMs, this knowledge also has limitations (as the decision tree’s performance is bounded) and therefore can be misleading. Thus, LLMs must use the predictive rule judiciously, especially when it fails. To examine this, we analyze 13 classification tasks from the TDC benchmark, whose test sets contain a total of 10498 molecules. Across these 13 classification tasks, the decision tree achieves an accuracy of 72.86\%, correctly predicting 7649 molecules. Figure \ref{fig-5} displays the distribution of the remaining 2849 molecules misclassified by the decision tree, grouped by whether they are correctly predicted by the LLMs. It can be seen that TreeKD still enables Gemma-2-2B and Granite-3.3-2B to correctly predict the majority, specifically 2241/2849 (78.66\%) and 2156/2849 (75.68\%), respectively. This demonstrates that LLMs do not blindly follow the predictive rule; instead, they can override it when the rule leads to incorrect predictions.

\begin{figure}
\centering
\includegraphics[width=1\linewidth]{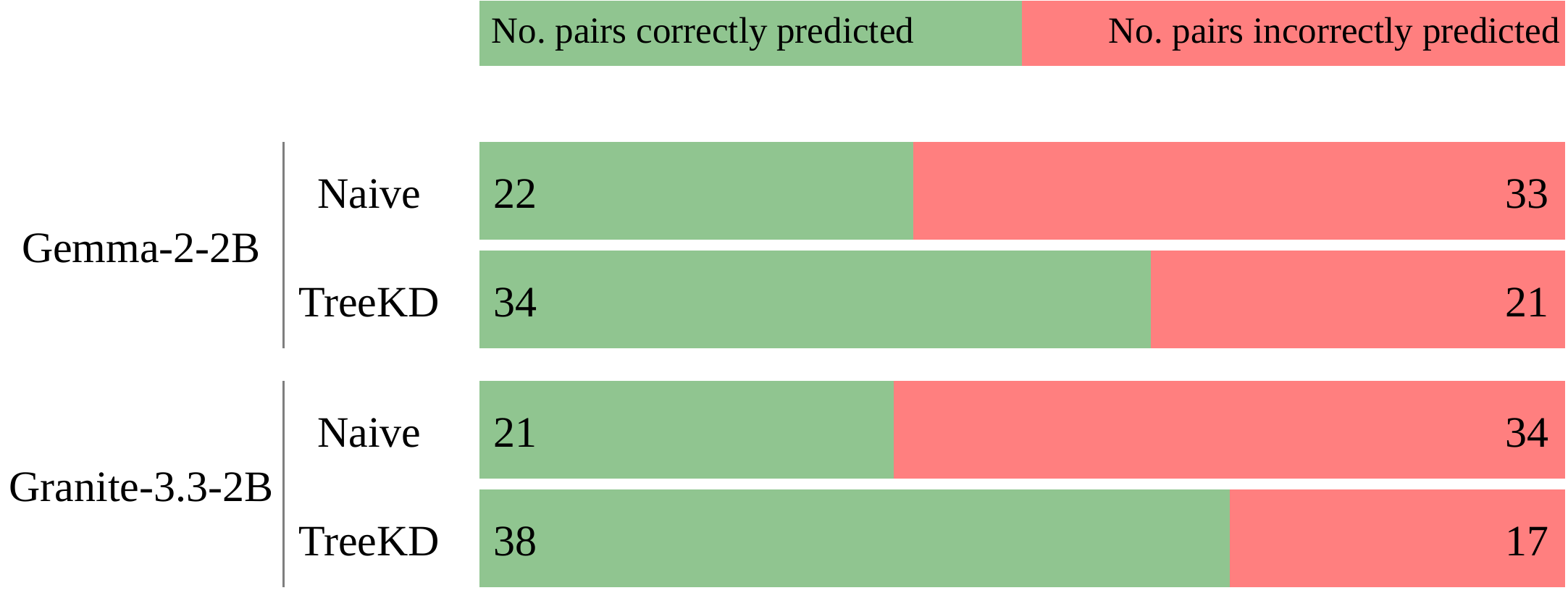}
\caption{
The distribution of 55 pairs of molecules exhibiting activity cliffs that the decision tree fails to distinguish, according to whether both molecules in each pair are correctly predicted by the LLMs. 
}
\label{fig-6}
\end{figure}

\begin{figure}
\centering
\includegraphics[width=1\linewidth]{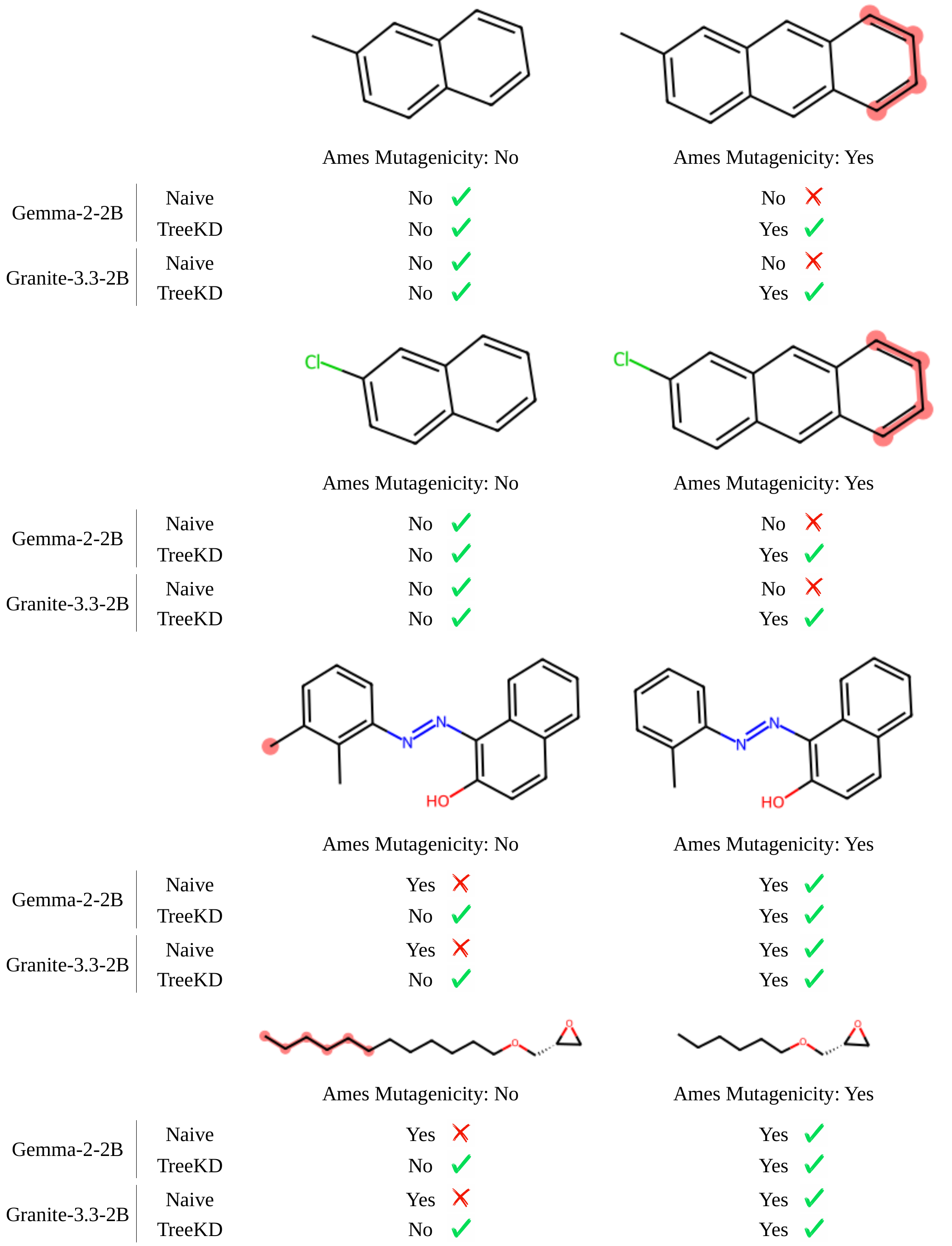}
\caption{
A few examples of pairs of molecules exhibiting activity cliffs on Ames Mutagenicity (differences between them are highlighted in \colorbox{myR}{red}), showing whether both molecules in each pair are correctly predicted. 
}
\label{fig-7}
\end{figure}

Furthermore, we hypothesize that TreeKD improves LLMs’ ability to capture the concept of activity cliffs. To examine this, we further analyze 13 classification tasks from the TDC benchmark. First, from the 10498 molecules, we identify 97 pairs of molecules exhibiting activity cliffs, where each pair consists of two molecules with a Tanimoto similarity \cite{bajusz2015tanimoto} greater than 0.8 but with opposite labels. Among these, we focus on the 55 pairs that the decision tree fails to distinguish. Figure \ref{fig-6} displays the distribution of these pairs, grouped by whether both molecules in each pair are correctly predicted by the LLMs. It can be seen that TreeKD enables Gemma-2-2B and Granite-3.3-2B to correctly predict both molecules in considerably more pairs (34/55 and 38/55) than the baseline (22/55 and 21/55). This suggests that TreeKD improves LLMs’ ability to capture the concept of activity cliffs, and further demonstrates that LLMs use the predictive rule wisely. Figure \ref{fig-7} shows a few examples on Ames Mutagenicity.

\section{Inference Cost}

\begin{figure}[!t]
\centering

\begin{subfigure}{\linewidth}
\centering
\includegraphics[width=\linewidth]{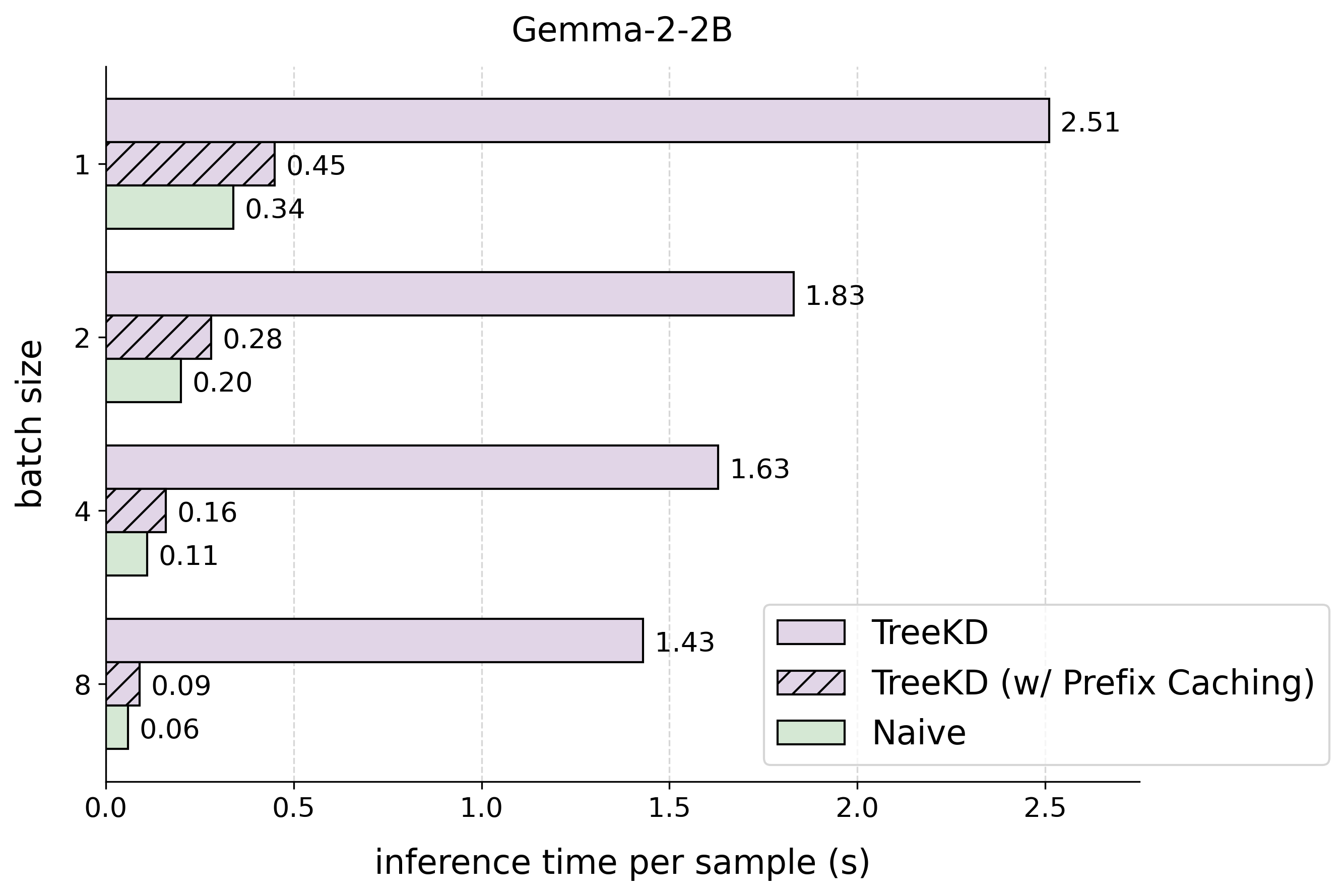}
\end{subfigure}

\vspace{0.1cm}

\begin{subfigure}{\linewidth}
\centering
\includegraphics[width=\linewidth]{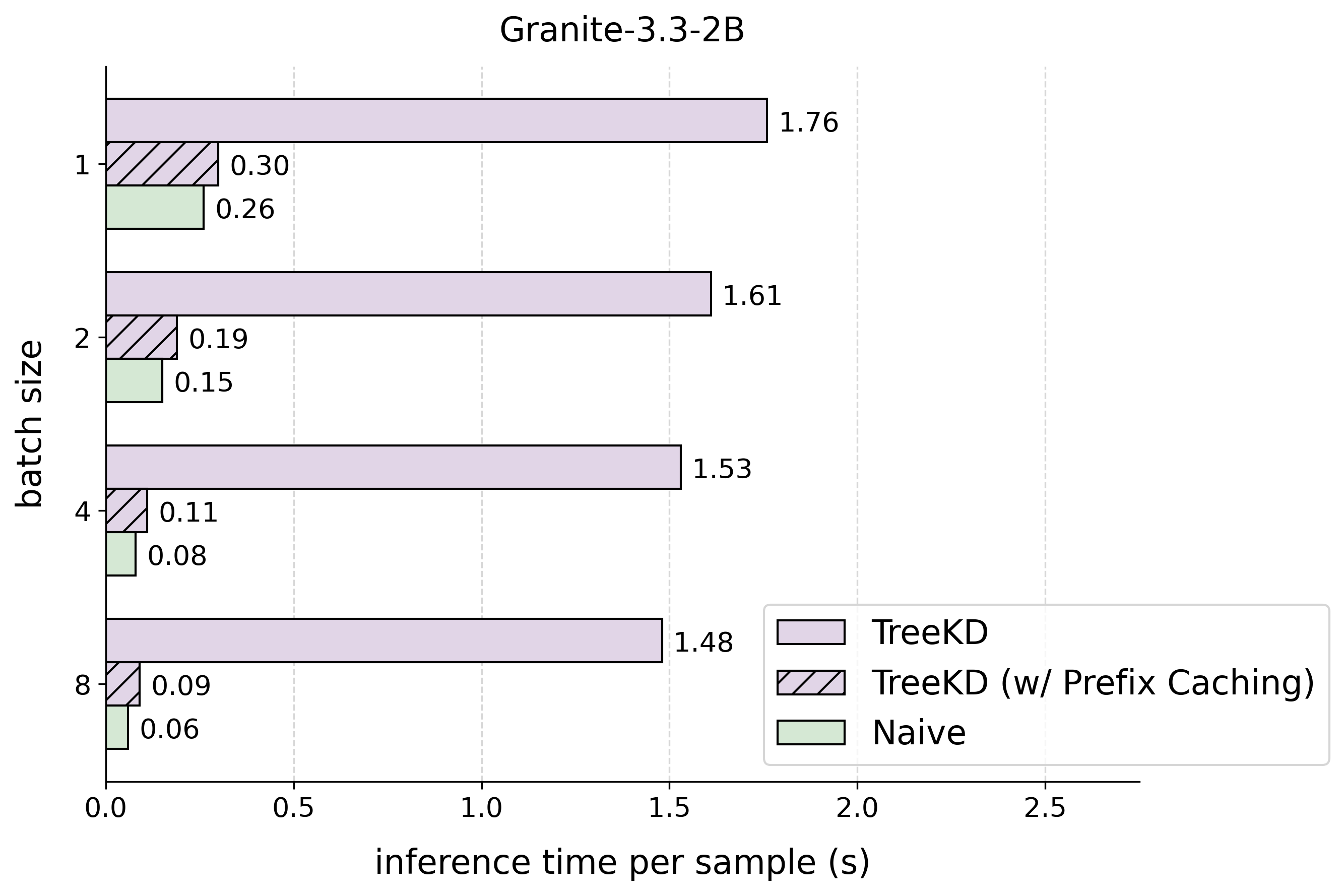}
\end{subfigure}

\caption{
The inference times (s) across different batch sizes. 
}
\label{fig-8}
\end{figure}

Prepending the predictive rule to the prompt increases TreeKD’s inference cost. However, this overhead is amortized as batch size increases during batch inference. Figure \ref{fig-8} shows inference times (in seconds) for different batch sizes. At a batch size of 8, TreeKD requires 1.43 (s) and 1.48 (s) with Gemma-2-2B and Granite-3.3-2B, respectively. Since the predictive rule is a static prefix, prefix-caching techniques \cite{gim2024prompt, zheng2024sglang} can further reduce latency. With prefix caching, the inference times decrease to 0.09 (s) for both Gemma-2-2B and Granite-3.3-2B, making them nearly identical to the 0.06 (s) required for naive inference. 

\section{Conclusion}
In this paper, we have presented TreeKD, a knowledge distillation method that transfers knowledge from tree-based specialist models to LLMs by verbalizing the learned predictive rule into natural language. We further introduced rule-consistency, a test-time scaling technique that aggregates predictions generated from different prompts constructed with different rules. Evaluation on the TDC benchmark with 22 prediction tasks shows that our method substantially enhances the performance of LLMs, outperforming the strongest generalist model and narrowing the gap with state-of-the-art specialist models. Our work suggests that combining traditional specialist models and LLMs offers a promising direction for advancing generalist models for MPP.

\section*{Limitations and Ethical Considerations}
One limitation of our proposed method is the increased training cost caused by longer prompts induced by the predictive rule. However, with an efficient training recipe and optimized frameworks such as vLLM \cite{kwon2023efficient}, this cost remains manageable, enabling all experiments to run on two NVIDIA A100 80GB GPUs. During inference, because the predictive rule is a static prefix, caching techniques \cite{gim2024prompt, zheng2024sglang} can be applied to reduce the cost of batch inference. In the future, expanding the current scope to include tens or hundreds of tasks, as well as generalizing the proposed method to other problems within the drug discovery pipeline and to additional entities such as proteins or nucleic acids \cite{ferreira2025ai}, constitutes an important direction. 

While we do not foresee negative societal impacts from our work, we acknowledge the potential for its dual use for nefarious purposes, such as the creation of hazardous molecules. We encourage discussions around these issues and strongly support the development and deployment of safeguards to prevent them.

\section*{Generative AI Usage}
The authors used a generative AI tool solely to improve the grammar and readability of portions of the manuscript. The AI-generated suggestions were verified and edited by the authors. The authors take full responsibility for the final content of the manuscript.

\bibliographystyle{ACM-Reference-Format}
\bibliography{sample-base}

\appendix

\begin{table}[!t]
\centering\footnotesize\renewcommand{\arraystretch}{1.1079}\setlength\tabcolsep{0pt}
\caption{
Detailed information and statistics of datasets. 
}
\label{tab-4}
\begin{tabular*}{\linewidth}{@{\extracolsep{\fill}} llrr }
\midrule
Type & Property & Dataset Size &Dataset Split \\
\midrule
\multirow[t]{9}{*}{Reg} 
&Caco-2 Permeability   &906   &Scaffold \\
&Lipophilicity         &4200  &Scaffold \\
&Solubility            &9982  &Scaffold \\
&PPBR                  &1797  &Scaffold \\
&VDss                  &1130  &Scaffold \\
&Half Life             &667   &Scaffold \\
&Clearance Microsome   &1102  &Scaffold \\
&Clearance Hepatocyte  &1020  &Scaffold \\
&LD50                  &7385  &Scaffold \\
\midrule
\multirow[t]{13}{*}{Cls}
&HIA                   &578   &Scaffold \\
&Pgp Inhibition        &1212  &Scaffold \\
&Bioavailability       &640   &Scaffold \\
&BBB                   &1975  &Scaffold \\
&CYP2D6 Inhibition     &13130 &Scaffold \\
&CYP3A4 Inhibition     &12328 &Scaffold \\
&CYP2C9 Inhibition     &12092 &Scaffold \\
&CYP2D6 Substrate      &664   &Scaffold \\
&CYP3A4 Substrate      &667   &Scaffold \\
&CYP2C9 Substrate      &666   &Scaffold \\
&hERG                  &648   &Scaffold \\
&Ames Mutagenicity     &7255  &Scaffold \\
&DILI                  &475   &Scaffold \\
\midrule
\end{tabular*}
\end{table}

\section{Hyperparameters of Specialist Models}
\label{app-a}
For each property $p \in P$, the decision tree $r_p$ as well as decision trees in the Random Forest are trained on the training dataset $D_p^{FGs} = \{(v_i, y_i)\}_{i=1}^{n_p}$. Every decision tree has a maximum depth of 6 (\verb|max_depth| = 6). Also, the minimum number of samples required to split an internal node is set to 2 (\verb|min_samples_split| = 2), and the minimum number of samples required to be at a leaf node is set to 1 (\verb|min_samples_leaf| = 1). All other hyperparameters are set to their default values as implemented in \texttt{scikit-learn} \cite{pedregosa2011scikit}.

\section{Training Details of LLMs}
\label{app-b}
Both Gemma-2-2B and Granite-3.3-2B are trained on the joint training dataset using Low-Rank Adaptation (LoRA) \cite{hu2022lora}. LoRA adapters are applied to all Linear modules except the generation head, with \verb|lora_rank| set to 16 and \verb|lora_alpha| set to 32. The LLMs are trained for 10 epochs using the Adam optimizer \cite{kingma2017adam} with \verb|weight_decay| set to 0 and an initial \verb|learning_rate| of 1e-5, which is linearly decayed to 1e-6 throughout training. Notably, prompts are truncated to a maximum length of 1024 tokens. A single training run with a batch size of 8 on two NVIDIA A100 80GB GPUs takes approximately 20 hours.

\section{TDC Benchmark}
\label{app-c}
The TDC benchmark \cite{huang2021therapeutics} is a resource for accessing datasets to evaluate AI methods for drug discovery. Concentrating on MPP, we use 22 prediction tasks corresponding to 22 ADMET properties, including Caco-2 Permeability, HIA (Human Intestinal Absorption), Pgp (P-glycoprotein) Inhibition, Bioavailability, Lipophilicity, Solubility, BBB (Blood-Brain Barrier), PPBR (Plasma Protein Binding Rate), VDss (Volume of Distribution at steady state), CYP2D6 Inhibition, CYP3A4 Inhibition, CYP2C9 Inhibition, CYP2D6 Substrate, CYP3A4 Substrate, CYP2C9 Substrate, Half Life, Clearance Microsome, Clearance Hepatocyte, hERG (Human Ether-a-go-go-related Gene), Ames Mutagenicity, DILI (Drug Induced Liver Injury), LD50 (Lethal Dose 50\%). Descriptions of these properties are directly adopted from the TDC benchmark’s site\protect\footnotemark. The dataset's information and statistics for each task are detailed in Table \ref{tab-4}. \footnotetext{https://tdcommons.ai}

\end{document}